\documentclass[11pt,a4paper]{article}
\usepackage[hyperref]{acl2019}
\usepackage{times}
\usepackage{latexsym}

\usepackage{url}

\usepackage{amsmath, amssymb}
\usepackage{float}
\usepackage{array, booktabs}
\usepackage{comment}
\usepackage{algorithm}
\usepackage{algorithmic}
\usepackage{graphicx}

\makeatletter
\newcommand{\ALOOP}[1]{\ALC@it\algorithmicloop\ #1%
  \begin{ALC@loop}}
\newcommand{\ENDALOOP}{\end{ALC@loop}\ALC@it\algorithmicendloop}

\newcommand{\algorithmicbreak}{\textbf{break}}
\newcommand{\BREAK}{\STATE \algorithmicbreak}
\newcommand{\algorithmiccontinue}{\textbf{continue}}
\newcommand{\CONTINUE}{\STATE \algorithmiccontinue}
\makeatother

\DeclareMathOperator*{\argmax}{arg\,max}

\aclfinalcopy 

\title{Multi-Task Semantic Dependency Parsing with Policy Gradient for \\ Learning Easy-First Strategies}

\author{Shuhei Kurita \\
  Center for Advanced Intelligence Project \\
  RIKEN \\
  Tokyo, Japan \\
  \texttt{shuhei.kurita@riken.jp} \\\And
  Anders S{\o}gaard \\
  Department of Computer Science \\
  University of Copenhagen \\
  Copenhagen, Denmark \\
  \texttt{soegaard@di.ku.dk} \\}

\date{}

\begin{document}
\maketitle

\begin{abstract}

In Semantic Dependency Parsing (SDP), semantic relations form directed acyclic graphs, rather than trees. We propose a new iterative predicate selection (IPS) algorithm for SDP. Our IPS algorithm combines the graph-based and transition-based parsing approaches in order to handle {\em multiple} semantic head words. We train the IPS model using a combination of multi-task learning and task-specific policy gradient training. Trained this way, IPS achieves a new state of the art on the SemEval 2015 Task 18 datasets. Furthermore, we observe that policy gradient training learns an easy-first strategy.

\end{abstract}

\section{Introduction}
Dependency parsers assign syntactic structures to sentences in the form of trees. Semantic dependency parsing (SDP), first introduced in the SemEval 2014 shared task \cite{oepen-EtAl:2014:SemEval}, in contrast, is the task of assigning {\em semantic} structures in the form of directed acyclic graphs to sentences.
SDP graphs consist of binary semantic relations, connecting semantic predicates and their arguments.
A notable feature of SDP is that words can be the semantic arguments of multiple predicates.
For example, in the English sentence: ``The man went back and spoke to the desk clerk'' -- the word ``man'' is the subject of the two predicates ``went back'' and ``spoke''.
SDP formalisms typically express this by two directed arcs, from the two predicates to the argument.
This yields a directed acyclic graph that expresses various relations among words.
However, the fact that SDP structures are directed acyclic graphs means that we cannot apply standard dependency parsing algorithms to SDP.

Standard dependency parsing algorithms are often said to come in two flavors:
{transition-based} parsers score transitions between states, and gradually build up dependency graphs on the side. {Graph-based} parsers, in contrast, score all candidate edges directly and apply tree decoding algorithms for the resulting score table. The two types of parsing algorithms have different advantages \cite{McDonald:Nivre:07}, with transition-based parsers often having more problems with error propagation and, as a result, with long-distance dependencies. This paper presents 
a compromise between transition-based and graph-based parsing, called {\em iterative predicate selection} (IPS) -- inspired by head selection algorithms for dependency parsing \cite{zhang-cheng-lapata2017} -- and show that error propagation, for this algorithm, can be reduced by a combination of multi-task and reinforcement learning.

\begin{figure}[t]
	\hspace{0em}
    \begin{center}
	\includegraphics[scale=1,clip]{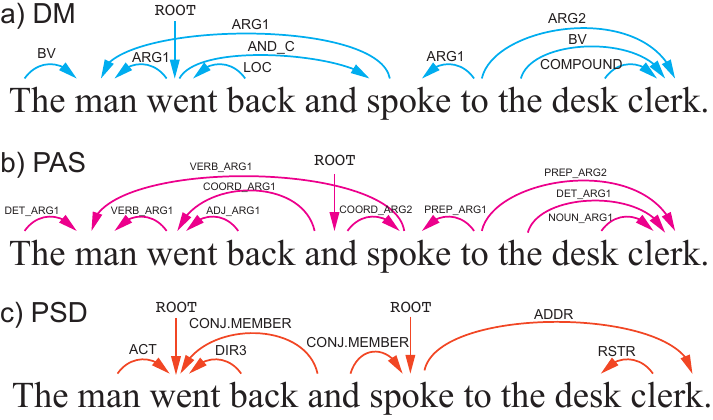}
	\vspace{0em}
       \caption{
			 Semantic dependency parsing arcs of DM, PAS and PSD formalisms.}
       \label{fig:parsing}
    \vspace{-1em}
    \end{center}
\end{figure}

Multi-task learning is motivated by the fact that there are several linguistic formalisms for SDP.
Fig.~\ref{fig:parsing} shows the three formalisms used in the shared task.
The DELPH-IN MRS (DM) formalism derives from DeepBank \cite{deepbank} and minimal recursion semantics \cite{copestake2005}.
Predicate-Argument Structure (PAS) is a formalism based on the Enju HPSG parser \cite{miyao2004} and is generally considered slightly more syntactic of nature than the other formalisms.
Prague Semantic Dependencies (PSD) are extracted from the Czech-English Dependency Treebank \cite{HAJI12.510.L12-1280}. There are several overlaps between these linguistic formalisms, and we show below that parsers, using multi-task learning strategies, can take advantage of these overlaps or synergies during training. 
Specifically, we follow \citeauthor{haopeng2017}~(2017) in using multi-task learning to learn representations of parser states that generalize better, but we go beyond their work, using a new parsing algorithm and showing that we can subsequently use reinforcement learning to prevent error propagation and tailor these representations to specific linguistic formalisms.

\paragraph{Contributions} In this paper,
(i) we propose a new parsing algorithm for semantic dependency parsing (SDP) that combines transition-based and graph-based approaches;
(ii) we show that multi-task learning of state representations for this parsing algorithm is superior to single-task training;
(iii) we improve this model by task-specific policy gradient fine-tuning; (iv) we achieve a new state of the art result across three linguistic formalisms; finally, (v) we show that policy gradient fine-tuning learns an easy-first strategy, which reduces error propagation.

\begin{figure*}[t]
	\centering
	\hspace{-1.0em}
	\includegraphics[scale=0.9,clip]{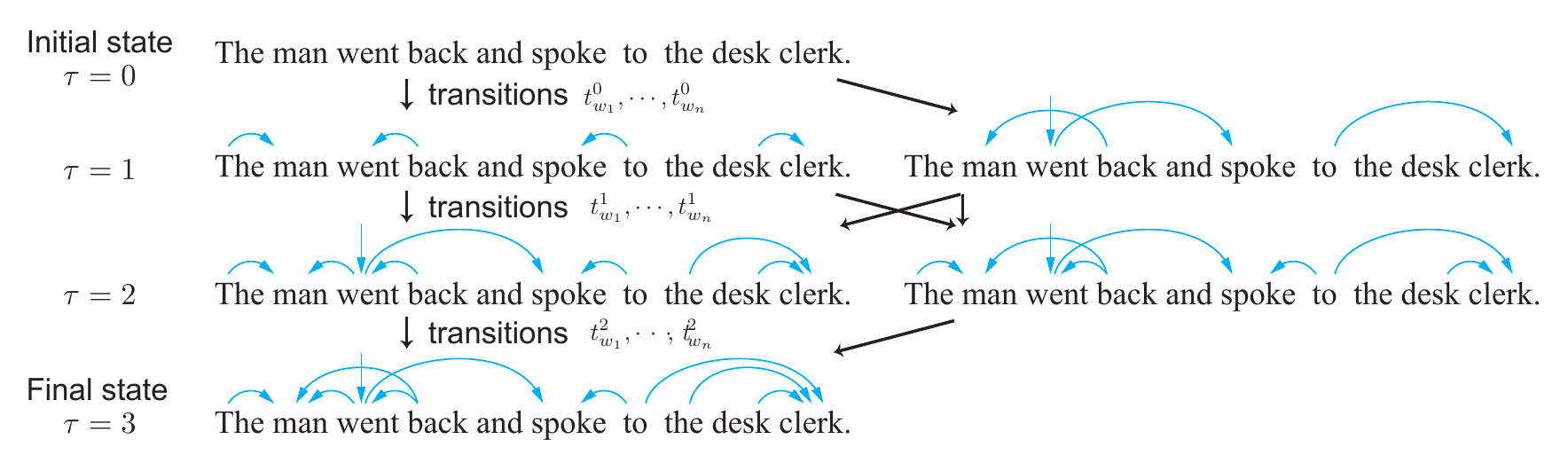}
       \caption{
Construction of semantic dependency arcs (DM) in the IPS parsing algorithm.
Parsing begins from the initial state and proceeds to the final state following one of several paths.
In the left path, the model resolves adjacent arcs first.
In contrast, in the right path, distant arcs that rely on the global structure are resolved first.
       }
       \label{fig:parsing_state_trans}
\end{figure*}
\section{Related Work}
There are generally two kinds of dependency parsing algorithms, namely transition-based parsing algorithms \cite{McDonald:Nivre:07, kiperwasser2016, ballesteros2015} and graph-based ones~\cite{mcdonald2006, zhang-clark2008:EMNLP, galley2009, zhang-cheng-lapata2017}. In graph-based parsing, a model is trained to score all possible dependency arcs between words, and decoding algorithms are subsequently applied to find the most likely dependency graph.
The Eisner algorithm \cite{eisner1996} and the Chu-Liu-Edmonds algorithm are often used for finding the most likely dependency trees, whereas the $AD^3$ algorithm \cite{martins-EtAl:2011:EMNLP1} is used for finding SDP graphs that form DAGs in \citeauthor{haopeng2017}~(2017) and \citeauthor{peng-etal-2018-learning}~(2018). 
During training, the loss is computed after decoding, leading the models to reflect a structured loss.
The advantage of graph-based algorithms is that there is no real error propagation to the extent the decoding algorithms are global inference algorithm, but this also means that reinforcement learning is not obviously applicable to graph-based parsing. 
In transition-based parsing, the model is typically taught to follow a gold transition path to obtain a perfect dependency graph during training. 
This training paradigm has the limitation that the model only ever gets to see states that are on gold transition paths, and error propagation is therefore likely to happen when the parser predicts wrong transitions leading to unseen states 
 \cite{McDonald:Nivre:07,goldberg2013}.

There have been several attempts to train transition-based parsers with reinforcement learning: \citeauthor{zhangchan2009}~(2009) 
applied SARSA \cite{leemon93} to an Arc-Standard model, using SARSA updates to fine-tune a model that was pre-trained using a feed-forward neural network. \citeauthor{daniel2018}~(2018), more recently, presented experiments with applying policy gradient training to several constituency parsers, including the RNNG transition-based parser \cite{dyer2016}.
In their experiments, however, the models trained with policy gradient did not always perform better than the models trained with supervised learning.
We hypothesize this is due to credit assignment being difficult in transition-based parsing. 
Iterative refinement approaches have been proposed in the context of sentence generation \cite{jason2018}.
Our proposed model explores multiple transition paths at once and avoids making risky decisions in the initial transitions, in part inspired by such iterative refinement techniques. We also pre-train our model with supervised learning to avoid sampling from irrelevant states at the early stages of policy gradient training. 





Several models have been presented for DAG parsing \cite{sagae2008,ribeyre-villemontedelaclergerie-seddah:2014:SemEval,alper2015,hershcovich2017}.
\citeauthor{wang2018}~(2018) proposed a similar transition-based parsing model for SDP;
they modified the possible transitions of the Arc-Eager algorithm \cite{nivre2004arceager} to create multi-headed graphs. We are, to the best of our knowledge, first to explore reinforcement learning for DAG parsing. 
\section{Model}

\subsection{Iterative Predicate Selection}
We propose a new semantic dependency parsing algorithm based on the head-selection algorithm for syntactic dependency parsing \cite{zhang-cheng-lapata2017}.
Head selection iterates over sentences, fixing the head of a word $w$ in each iteration, ignoring $w$ in future iterations.
This is possible for dependency parsing because each word has a unique head word, including the root of the sentence, which is attached to an artificial root symbol.
However, in SDP, words may attach to multiple head-words or {\it semantic predicates} whereas other words may not attach to any semantic predicates.
Thus, we propose an iterative predicate selection (IPS) parsing algorithm, as a generalization of head-selection in SDP.

The proposed algorithm is formalized as follows. 
First, we define transition operations for all words in a sentence.
For the $i$-th word $w_i$ in a sentence, the model selects one transition $t^{\tau}_i$ from the set of possible transitions $T^{\tau}_i$
for each transition time step $\tau$. Generally, the possible transitions $T_i$ for the $i$-th word are expressed as follows:
$$
\{\mathrm{NULL},\mathrm{ARC}_{i,\mathrm{ROOT}},\mathrm{ARC}_{i,1},\cdots,\mathrm{ARC}_{i,n}\}
$$
where $\mathrm{ARC}_{i,j}$ is a transition to create an arc from the $j$-th word to the $i$-th word, encoding that the semantic predicate $w_j$ takes $w_i$ as an semantic argument.
{\usefont{T1}{pcr}{m}{n} NULL} is a special transition that does not create an arc.
The set of possible transitions $T_{i}^{\tau}$ for the $i$-th word at time step $\tau$ is a subset of possible transitions $T_i$ that satisfy two constraints: (i) no arcs can be reflexive, i.e.,  $w_i$ cannot be an argument of itself, and (ii) the new arc must not be a member of the set of arcs $A^{\tau}$ comprising the partial parse graph $\mathbf{y}^{\tau}$ constructed at time step $\tau$.
Therefore, we obtain: $T_i^{\tau}=T_i/(\mathrm{ARC}_{i,i}\cup A^{\tau})$.
The model then creates semantic dependency arcs by iterating over the sentence as follows:\footnote{This algorithm can introduce circles. However, circles were extremely rare in our experiments, and can be avoided by simple heuristics during decoding. We discuss this issue in the Supplementary Material, \S\ref{sec:dag_stat}.}
\begin{description}
	\item[1] For each word $w_i$, select a head arc from $T^{\tau}_i$.
    \item[2] Update the partial semantic dependency graph.
	\item[3] If all words select {\usefont{T1}{pcr}{m}{n} NULL}, the parser halts. Otherwise, go to {\bf 1}.
\end{description}

%

Fig.~\ref{fig:parsing_state_trans} shows the transitions of the IPS algorithm during the DM parsing of the sentence ``The man went back and spoke to the desk clerk.'' 
In this case, there are several paths from the initial state to the final parsing state,
depending on the orders of creating the arcs.
This is known as the non-deterministic oracle problem \cite{goldberg2013}.
In IPS parsing, some arcs are easy to predict; others are very hard to predict. 
Long-distance arcs are generally difficult to predict, but they are very important for down-stream applications, including reordering for machine translation \cite{Xu:ea:09}.
Since long-distance arcs are harder to predict, and transition-based parsers are prone to error propagation, several easy-first strategies have been introduced, both in supervised \cite{goldberg2010,ma-etal-2013-easy} and unsupervised dependency parsing \cite{spitkovsky2011}, to prefer some paths over others in the face of the non-deterministic oracle problem.
Easy-first principles have also proven effective with sequence taggers \cite{tsuruoka-tsujii-2005-bidirectional,martins-kreutzer-2017-learning}. 
In this paper, we take an arguably more principled approach, {\em learning}~a strategy for choosing transition paths over others using reinforcement learning. We observe, however, that the learned strategies exhibit a clear easy-first preference. 

\begin{figure*}[t]
  \centering
	\includegraphics[scale=0.75,clip]{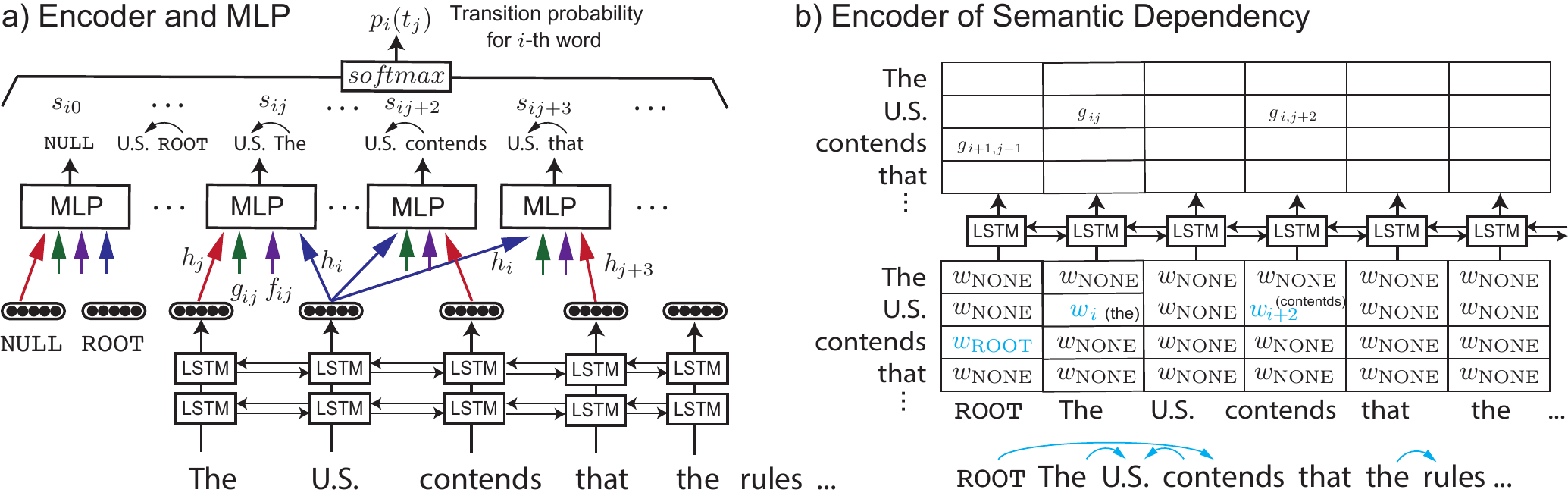}
       \caption{Our network architecture: (a) The encoder of the sentence into the hidden representations  $h_i$ and $h_j$, and the MLP for the transition probabilities. (b) The encoder of the semantic dependency matrix for the representation of $h^d_{ij}$. The MLP also takes the arc flag representation $f_{ij}$ (see text for explanation).
       }
       \label{fig:neural_net_main}
    \vspace{-0.5em}
\end{figure*}

\subsection{Neural Model}
Fig.~\ref{fig:neural_net_main} shows the overall neural network.
It consists of an encoder for input sentences and partial SDP graphs, as well as a multi-layered perceptron (MLP) for the semantic head-selection of each word.


\paragraph{Sentence encoder}
We employ bidirectional long short-term memory (BiLSTM) layers for encoding words in sentences.
A BiLSTM consists of two LSTMs that reads the sentence forward and backward, and concatenates their output before passing it on.
For a sequence of tokens $[w_1, \cdots, w_n]$, the inputs for the encoder are words, POS tags and lemmas.\footnote{In the analysis of our experiments, we include an ablation test, where we leave out lemma information for a more direct comparison with one of our baselines.}
They are mapped to the same $p$-dimensional embedding vectors in a look-up table. Then they are concatenated
to form $3p$-dimensional vectors and used as the input of BiLSTMs.
We denote the mapping function of tokens into $3p$-dimensional vectors as $u(w_{*})$ for later usages.
Finally, we obtain the hidden representations of all words $[h(w_1),\cdots,h(w_n)]$ from the three-layer BiLSTMs.
We use three-layer stacked BiLSTMs.
We also use special embeddings $h_{\mathrm{NULL}}$ for the {\usefont{T1}{pcr}{m}{n} NULL} transition and $h_{\mathrm{ROOT}}$ for the {\usefont{T1}{pcr}{m}{n} ROOT} of the sentence.

\paragraph{Encoder of partial SDP graphs}
The model updates the partial SDP graph at each time step of the parsing procedure.
The SDP graph $\mathbf{y}^{\tau}$ at time step $\tau$ is stored in a semantic dependency matrix $G^{\tau} \in \{0,1\}^{n \times (n+1)}$ for a sentence of $n$ words.\footnote{In this subsection, we omit the time step subscription $\tau$ of the partial SDP graph from some equations for simplicity.}
The rows of the matrix $G$ represent arguments and the columns represent head-candidates, including the {\usefont{T1}{pcr}{m}{n} ROOT} of the sentence, which is represented by the first column of the matrix. For each transition for a word, the model fills in one cell in a row, if the transition is not {\usefont{T1}{pcr}{m}{n} NULL}.
In the initial state, all cells in $G$ are 0.
A cell $G[i,j]$ is updated to 1, when the model predicts that the $(i-1)$-th word is an argument of the $j$-th word or {\usefont{T1}{pcr}{m}{n} ROOT} when $j=0$.



We convert the semantic dependency matrix $G$ into
a rank three tensor $G' \in \mathbb{R}^{n \times (n+1) \times p}$,
by replacing elements with embeddings of tokens $u(w_{*})$ by
\begin{equation}
 g'_{ij} = \begin{cases}
 u(w_{j-1}) & (g_{ij}=1)\\
 u(w_{\mathrm{NONE}}) & (g_{ij}=0)
\end{cases}
\end{equation} where $g_{ij} \in G$ and $g'_{ij} \in G'$.
$g'_{i*}$ contains the representations of the semantic predicates for the $i$-th word in the partial SDP graph.
We use a single layer Bi-LSTM to encode the semantic predicates $g'_{i*}$ of each word; see Fig. \ref{fig:neural_net_main} (b).
Finally, we concatenate the hidden representation of the {\usefont{T1}{pcr}{m}{n} NULL} transition and obtain the partial SDP graph representation $G^{\tau}$ of the time step $\tau$:
\begin{equation}
G^{\tau} = [g^{\tau}_{\mathrm{NULL}},g^{\tau}_{*,1}, \cdots, g^{\tau}_{*,n+1}]
\end{equation}

We also employ dependency flags that directly encode the semantic dependency matrix and indicate whether the corresponding arcs are already created or not. Flag representations $F'$ are also three-rank tensors, consisting of two hidden representations: $f_\mathrm{ARC}$ for $g_{i,j}=1$ and $f_\mathrm{NOARC}$ for $g_{i,j}=0$ depending on $G$.  
$f_\mathrm{ARC}$ and $f_\mathrm{NOARC}$ is $q$-dimensional vectors.
Then we concatenate the hidden representation of the {\usefont{T1}{pcr}{m}{n} NULL} transition and obtain the flag representation $F^{\tau}$:
\begin{equation}
F^{\tau} = [f^{\tau}_{\mathrm{NULL}},f^{\tau}_{*,1},\cdots,f^{\tau}_{*,n+1}]
\end{equation}.
We do not use BiLSTMs to encode these flags. These flags also reflect the current state of the semantic dependency matrix.


\paragraph{Predicate selection model}
The semantic predicate selection model comprises an MLP with inputs from the encoder of the sentence and the partial semantic dependency graph: the sentence representation $H$, the SDP representation $G^{\tau}$, and the dependency flag $F^{\tau}$.
They are rank three tensors and concatenated at the third axis.
Formally, the score $s_{ij}$ of the $i$-th word and the $j$-th transition is expressed as follows.
\begin{equation}
s^{\tau}_{ij} = \mathrm{MLP}([h_i,h_j,g^{\tau}_{ij},f^{\tau}_{ij}])
\end{equation}
For the MLP, we use a concatenation of outputs from three different networks: a three-layer MLP, a two-layer MLP and a matrix multiplication with bias terms
as follows.
\begin{eqnarray*}
	\mathrm{MLP}(\mathbf{x}) = W^3_3 a\big(W^3_2 a(W^3_1 \mathbf{x}+b^3_1)+b^3_2\big) \\
	+ W^2_2 a(W^2_1 \mathbf{x}+b^2_2) + W^1_1 \mathbf{x}+b^1_1
	\label{eq:fnn}
\end{eqnarray*}
$W^{*}_{*'}$ are matrices or vectors used in this MLP and $W^{*}_{*'}$ are bias terms.
Here, we use this MLP for predicting a scalar score $s_{ij}$; therefore, $W^3_3,W^2_2,W^1_1$ are vectors.
The model computes the probability of the transition $t_j$ for each word $i$ by applying a softmax function over the candidates of the semantic head words $w_j$.
\begin{equation}
	p_{i}(t^{\tau}_j) = \mathrm{softmax}_j(s^{\tau}_{ij})
\end{equation}
These transition probabilities $p_{i}(t_j)$ of selecting a semantic head word $w_j$, are defined for each word $w_i$ in a sentence.

For supervised learning, we employ a cross entropy loss
\begin{equation}
L^{\tau}(\theta) = - \sum_{i,j} l_{i} \log p_i(t_j^{\tau}|G^{\tau})
\label{eq:sp}
\end{equation} for the partial SDP graph $G^{\tau}$ at time step $\tau$.
Here $l_{i}$ is a gold transition label for the $i$-th word
and $\theta$ represents all trainable parameters. 
Note that this supervised training regime, as mentioned above, does not have a principled answer to the non-deterministic oracle problem \cite{goldberg2013}, and samples transition paths randomly from those consistent with the gold anntoations to create transition labels.

\paragraph{Labeling model}
We also develop a semantic dependency labeling neural network.
This neural network consists of three-layer stacked BiLSTMs and
a MLP for predicting a semantic dependency label between words and their predicates.
We use a MLP that is a sum of the outputs from a three-layer MLP, a two-layer MLP and a matrix multiplication.
Note that the output dimension of this MLP is the number of semantic dependency labels.
The input of this MLP is the hidden representations of a word $i$ and its predicates $j$: $[h_i,h_j]$ extracted from the stacked BiLSTMs.
The score $s'_{ij}(l)$ of the label $l$ for the arc from predicate $j$ to word $i$ is predicted as follows.
\begin{equation}
	s'_{ij}(l) = \mathrm{MLP'}([h_i,h_j])
\end{equation}
We minimize the softmax cross entropy loss using supervised learning.

\subsection{Reinforcement Learning}

\begin{algorithm}[t]
\caption{\small{Policy gradient learning for IPS Algorithm}}
\label{alg:pg}
\small\begin{algorithmic}
\REQUIRE Sentence $\mathbf{x}$ with an empty parsing tree $\mathbf{y}^0$.
\STATE Let a time step $\tau=0$ and finish flags $f_*=0$.
\FOR{ $0 \leq \tau <$ the number of maximum iterations }
\STATE Compute $\pi^{\tau}$ and argmax transitions $\hat{t}_i=\argmax{\pi_i^{\tau}}$.
\IF{$\forall i$\ ;\ $\hat{t}_i^{\tau} = \mathrm{NULL} $}
 \BREAK
\ENDIF
\FOR{ $i$-th word in a sentence }
\IF{check a finish flag $f_i=1$}
	\CONTINUE
\ENDIF
 \IF{all arcs to word $i$ are correctly created in $\mathbf{y^{\tau}}$ and $\hat{t}_i=\mathrm{NULL}$}
 	\STATE Let a flag $f=1$
   \CONTINUE
 \ENDIF
 \STATE Sample $t_i^{\tau}$ from $\pi_i^{\tau}$.
 \STATE Update the parsing tree $\mathbf{y}^{\tau}$ to $\mathbf{y}^{\tau+1}$.
 \STATE Compute a new reward $r_i^{\tau}$ from $\mathbf{y}^{\tau}$, $\mathbf{y}^{\tau+1}$ and $\mathbf{y}^{g}$.
\ENDFOR
 \STATE Store a tuple of the state, transitions and rewards for words $\{\mathbf{y}^{\tau}, t_*^{\tau},r_*^{\tau}\}$.
\ENDFOR
\STATE Shuffle tuples of $\{\mathbf{y}^{\tau}, t_*^{\tau},r_*^{\tau}\}$ for a time step $\tau$.
\FOR{a tuple $\{\mathbf{y}^{\tau'}, t_*^{\tau},r_*^{\tau'}\}$ of time step $\tau'$ }
\STATE Compute gradient and update parameters.
\ENDFOR
\end{algorithmic}
\end{algorithm}

\paragraph{Policy gradient} Reinforcement learning is a method for learning to iteratively act according to a dynamic environment in order to optimize future rewards.
In our context, the agent corresponds to the neural network model predicting the transition probabilities $p_i(t^{\tau}_j)$ that are used in the parsing algorithm.
The environment includes the partial SDP graph $\mathbf{y}^{\tau}$, and the rewards $r^{\tau}$ are computed by comparing the predicted parse graph to the gold parse graph $\mathbf{y}^g$.

We adapt a variation of the policy gradient method \cite{policygradient1992} for IPS parsing.
Our objective function is to maximize the rewards
\begin{equation}
	J(\theta) = E_{\pi} \left[ r_{i}^{\tau}  \right]
	\label{eq:rl}
\end{equation}
and the transition policy for the $i$-th word is given by the probability of the transitions
	$\pi \sim p_i(t_j^{\tau}|\mathbf{y}^{\tau})$.
The gradient of Eq.\ref{eq:rl} is given as follows:
\begin{equation}
	\nabla J(\theta) = E_{\pi} \left[ r_{i}^{\tau} \nabla \log p_i(t_j^{\tau}|\mathbf{y}^{\tau}) \right]
\end{equation}
When we compute this gradient, given a policy $\pi$,
we approximate the expectation $E_{\pi}$ for any transition sequence with a single transition path $\mathbf{t}$ that is sampled from policy $\pi$:
\begin{equation}
	\nabla J(\theta) \approx \sum_{t_j^{\tau} \in \mathbf{t}} \large[ r_{i}^{\tau}  \nabla \log p_i(t_j^{\tau}|\mathbf{y}^{\tau}) \large]
\end{equation}

We summarize our policy gradient learning algorithm for SDP in Algorithm \ref{alg:pg}.
For time step $\tau$, the model samples one transition $t_j^{\tau}$ selecting the $j$-th word as a semantic head word of the $i$-th word, from the set of possible transitions $T_i$, following the transition probability of $\pi$.
After sampling $t_j^{\tau}$, the model updates the SDP graph to $\mathbf{y}^{\tau+1}$ and computes the reward $r_{i}^{\tau}$.
When {\usefont{T1}{pcr}{m}{n} NULL} becomes the most likely transition for all words, or the time step exceeds the maximum number of time steps allowed, we stop.\footnote{We limit the number of transitions during training, but not at test time.}
For each time step, we then update the parameters of our model with the gradients computed from the sampled transitions and their rewards.\footnote{We update the parameters for each time step to reduce memory requirements.}

%

Note how the cross entropy loss and the policy gradient loss are similar, if we do not sample from the policy $\pi$, and rewards are non-negative. However, these are the important differences between supervised learning
and reinforcement learning:
(1) Reinforcement learning uses sampling of transitions. This allows our model to explore transition paths that supervised models would never follow.
(2) In supervised learning, decisions are independent of the current time step $\tau$, while in reinforcement learning, decisions depend on $\tau$.
This means that the $\theta$ parameters are updated {\em after} the parser finishes parsing the input sentence.
(3) Loss must be non-negative in supervised learning, while rewards can be negative in reinforcement learning.

In general, the cross entropy loss is able to optimize for choosing good transitions given a parser configuration, while the policy gradient objective function is able to optimize the entire sequence of transitions drawn according to the current policy.
We demonstrate the usefulness of reinforcement learning in our experiments below.

\begin{table}[t]
\begin{center}
\small\begin{tabular}{ll}
        \toprule
        Reward               & Transitions  \\
        \midrule
        $r^{\tau}_i=1$  & (1) The model creates a new correct arc from\\
                        &   a semantic predicate to the $i$-th word. \\
				                & (2) The first time the model chooses the {\usefont{T1}{pcr}{m}{n} NULL}\\
                        & transition after all gold arcs to the $i$-th word \\
                        & have been created, and no wrong arcs to the $i$ \\
                        & words have not been created. \\
				$r^{\tau}_i=-1$ & (3) The model creates a wrong arc from a \\
                        & semantic predicate candidate to the $i$-th word. \\
        $r^{\tau}_i=0$  & (4) All other transitions. \\
        \bottomrule
\end{tabular}
    \caption{Rewards in SDP policy gradient. }
    \label{table:rewards}
    \vspace{-1em}
\end{center}
\end{table}

\paragraph{Rewards for SDP} 
We also introduce intermediate rewards, given during parsing, at different time steps.
The reward $r^{\tau}_i$ of the $i$-th word is determined as shown in Table \ref{table:rewards}.
The model gets a positive reward for creating a new correct arc to the $i$-th word, or if the model for the first time chooses a {\usefont{T1}{pcr}{m}{n} NULL} transition after all arcs to the $i$-th word are
correctly created.
The model gets a negative reward when the model creates wrong arcs.
When our model chooses {\usefont{T1}{pcr}{m}{n} NULL} transitions for the $i$-th word before all gold arcs are created, the reward $r^{\tau}_i$ becomes 0.

\subsection{Implementation Details}
This section includes details of our implementation.\footnote{The code is available at \url{https://github.com/shuheikurita/semrl}}
We use 100-dimensional, pre-trained Glove \cite{pennington2014glove} word vectors.
Words or lemmas in the training corpora that do not appear in pre-trained embeddings are associated with randomly initialized vector representations.
Embeddings of POS tags and other special symbol are also randomly initialized.
We apply Adam as our optimizer.
Preliminary experiments show that mini-batching led to a degradation in performance. When we apply policy gradient, we pre-train our model using supervised learning. We then use policy gradient for task-specific fine-tuning of our model.
We find that updating parameters of BiLSTM and word embeddings during policy gradient makes training quite unstable. Therefore we fix the BiLSTM parameters during policy gradient.
In our multi-task learning set-up,
we apply multi-task learning of the shared stacked BiLSTMs \cite{ander2016,hashimoto2017} in supervised learning.
We use task-specific MLPs for the three different linguistic formalisms: DM, PAS and PSD. 
We train the shared BiLSTM using multi-task learning beforehand, and then we fine-tune the task-specific MLPs with policy gradient.
We summarize the rest of our hyper-parameters in Table \ref{table:hyperparams}.

\begin{table}[t]
\begin{center}
	\small\begin{tabular}{lc}
        \toprule
				Name & Value \\
				\midrule
				Encoder BiLSTM hidden layer size   & 600 \\
				Dependency LSTM hidden layer size & 200 \\
        The dimensions of embeddings $p$,$q$ & 100, 128 \\
        MLPs hidden layer size    & 4000 \\
				Dropout rate in MLPs & 0.5 \\
        Max transitions during reinforcement learning & 10 \\
        \bottomrule
	\end{tabular}
    \caption{
			Hyper-parameters in our experiments.
    }
    \label{table:hyperparams}
    \vspace{-1em}
\end{center}
\end{table}

\section{Experiments}

We use the SemEval 2015 Task18 \cite{oepen-EtAl:2015:SemEval} SDP dataset for evaluating our model.
The training corpus contains 33,964 sentences from the WSJ corpus; the development and in-domain test were taken from the same corpus and consist of 1,692 and 1,410 sentences, respectively.
The out-of-domain test set of 1,849 sentences is drawn from Brown corpus.
All sentences are annotated with three semantic formalisms: DM, PAS and PSD.
We use the standard splits of the datasets \cite{mariana2015,yantao2015}.
Following standard evaluation practice in semantic dependency parsing, all scores are {\em micro-averaged} F-measures \cite{haopeng2017,wang2018} with labeled attachment scores (LAS).

\begin{table}[t]
\begin{center}
	\small\begin{tabular}{lcccc}
        \toprule
        Model               & DM  & PAS  & PSD & Avg.  \\
        \midrule
        Peng+ 17~Freda3  & 90.4 & 92.7 & 78.5 & 88.0 \\
		    Wang+ 18~Ens. & 90.3 & 91.7	& 78.6 & 86.9 \\
        Peng+ 18~   & 91.6 & -    & 78.9 & -    \\
        \midrule
        IPS           & 91.1 & 92.4 & 78.6 & 88.2 \\
	    IPS +ML		  & 91.2 & 92.5 & 78.8 & 88.3 \\
	    IPS +RL		  & 91.6$^\ddagger$  & \textbf{92.8}$^\ddagger$  & 79.2$^\ddagger$  & 88.7$^\ddagger$  \\
	    IPS +ML +RL	  & \textbf{92.0}$^\ddagger$ & \textbf{92.8}$^\ddagger$ & \textbf{79.3}$^\ddagger$ & \textbf{88.8}$^\ddagger$ \\
        \bottomrule
	\end{tabular}
    \caption{
		Labeled parsing performance on in-domain test data.
		Avg. is the micro-averaged score of three formalisms.
		$\ddagger$ of the +{\em RL} models represents that the scores are statistically significant at $p<10^{-3}$ with their non-{\em RL} counterparts.
    }
    \vspace{-0.5em}
    \label{table:las}
\end{center}
\end{table}

\begin{table}[t]
\begin{center}
	\small\begin{tabular}{lcccc}
        \toprule
        Model               & DM  & PAS  & PSD & Avg.  \\
        \midrule
        Peng+ 17~Freda3  & 85.3 & \textbf{89.0} & 76.4 & 84.4 \\
        Peng+ 18~        & 86.7 & -    & 77.1 & -    \\
        \midrule
	  		IPS +ML		     & 86.0 & 88.2 & 77.2 & 84.6 \\
	      IPS +ML +RL	   & \textbf{87.2}$^\ddagger$ & 88.8$^\ddagger$ & \textbf{77.7}$^\ddagger$ & \textbf{85.3}$^\ddagger$ \\
        \bottomrule
	\end{tabular}
    \caption{
		Labeled parsing performance on out-of-domain test data.
		Avg. is the micro-averaged score of three formalisms.
		$\ddagger$ of the +{\em RL} models represents that the scores are statistically significant at $p<10^{-3}$ with their non-{\em RL} counterparts.
    }
    \vspace{-0.75em}
    \label{table:ood}
\end{center}
\end{table}

\begin{figure*}[t]
	\centering
	\includegraphics[scale=0.65,clip]{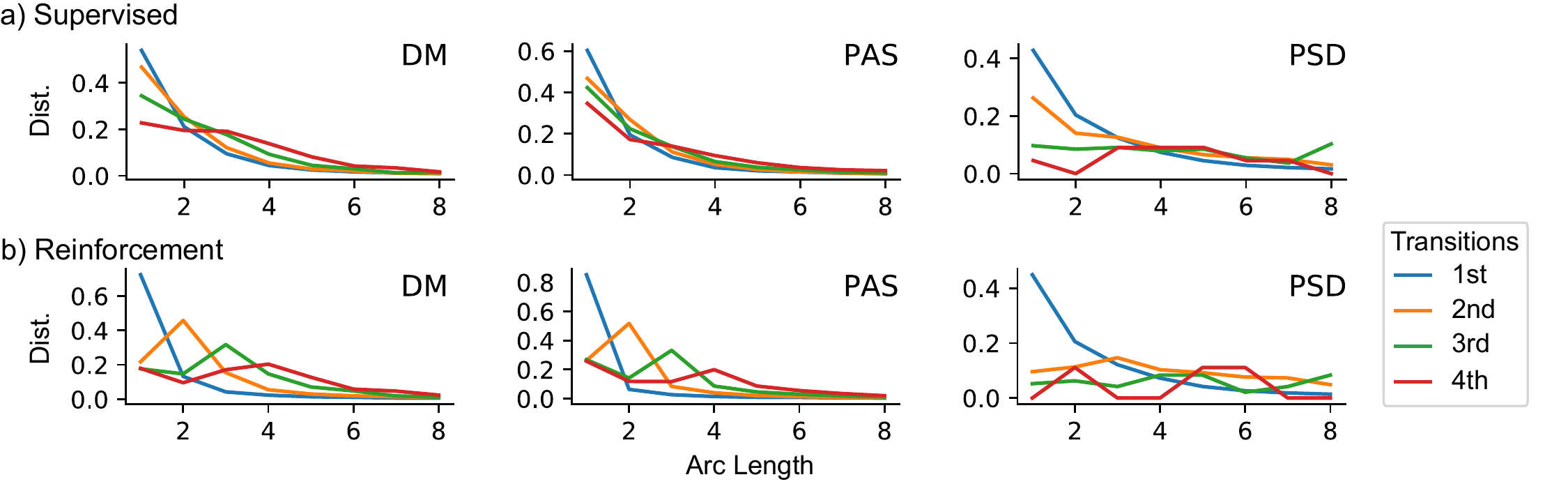}
	\vspace{-0.5em}
       \caption{
			 Arc length distributions:
			 (a) Supervised learning ({\em IPS+ML}).
			 (b) Reinforcement learning  ({\em IPS+ML+RL}).
			 The four lines correspond to the first to fourth transitions in the derivations.
       }
       \label{fig:distance}
\end{figure*}

\begin{figure*}[t]
	\centering
	\includegraphics[scale=0.95,clip]{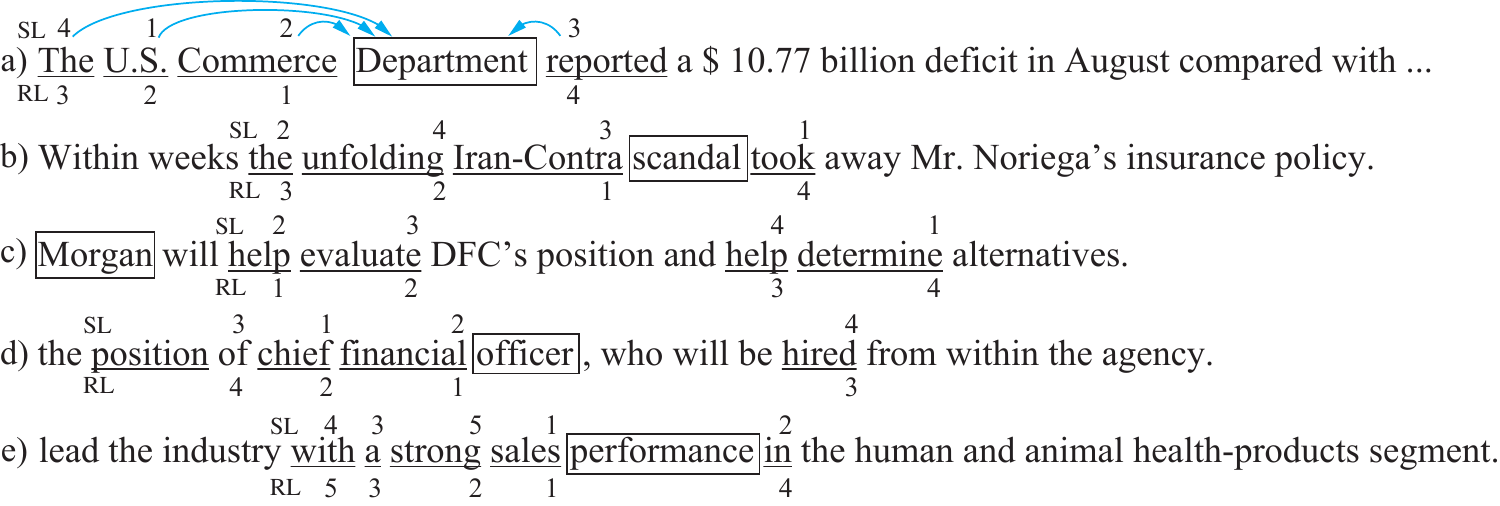}
       \caption{
			 Examples of clauses parsed with DM formalism.
			 The underlined words are the semantic predicates of the argument words in rectangles in the annotation.
			 The superscript numbers (SL) are the orders of creating arcs by {\em IPS+ML} and the subscript numbers (RL) are the orders by {\em IPS+ML+RL}.
			 In the clause (a), we show a partial SDP graph to visualize the SDP arcs.
       }
       \label{fig:examples}
\end{figure*}

The system we propose is the IPS parser trained with a multi-task objective and fine-tuned using reinforcement learning. This is referred to as {\em IPS+ML+RL} in the results tables. To highlight the contributions of the various components of our architecture, we also report ablation scores for the IPS parser without multi-task training nor reinforcement learning ({\em IPS}), with multi-task training ({\em IPS+ML}) and with reinforcement learning ({\em IPS+RL}).
At inference time, we apply heuristics to avoid predicting circles during decoding \cite{dag_form1980}; see Supplementary Material, \S\ref{sec:dag_stat}. This improves scores by 0.1 \% or less, since predicted circles are extremely rare. 
We compare our proposed system with three state-of-the-art SDP parsers: Freda3 of \citeauthor{haopeng2017} (2017), the ensemble model in \citeauthor{wang2018} (2018) and \citeauthor{haopeng2018b} (2018). In \citeauthor{haopeng2018b} (2018), they use syntactic dependency trees, while we do not use them in our models.\footnote{\citet{dozat2018} report {\em macro-averaged} scores instead, as mentioned in their ACL 2018 talk, and their results are therefore not comparable to ours.
For details, see the video of their talk on ACL2018 that is available on Vimeo.
}

The results of our experiments on in-domain dataset are also shown in Table \ref{table:las}.
We observe that our basic {\em IPS} model achieves competitive scores in DM and PAS parsing.
Multi-task learning of the shared BiLSTM ({\em IPS+ML}) leads to small improvements across the board, which is consistent with the results of \citeauthor{haopeng2017}~(2017). The model trained with reinforcement learning ({\em IPS+RL}) performs better than the model trained by supervised learning ({\em IPS}). These differences are significant ($p<10^{-3}$).
Most importantly, the combination of multi-task learning and policy gradient-based reinforcement learning ({\em IPS+ML+RL}) achieves the best results among all IPS models and the previous state of the art models, by some margin. We also obtain similar results for the out-of-domain datasets, as shown in Table~\ref{table:ood}. All improvements with reinforcement learning are also statistically significant ($p<10^{-3}$).

\begin{table}[t]
\begin{center}%
	\small\begin{tabular}{llccc}
       \toprule
       Model         & DM  & PAS  & PSD & Avg. \\
       \midrule
       Peng+ 17~Freda3     & 90.4 & 92.5 & 78.5 & 88.0 \\
			 \midrule
			 IPS +ML~~~~~~~~~-Lemma     & 90.7 & 92.3 & 78.3 & 88.0 \\
			 IPS +ML +RL~-Lemma & \textbf{91.2}$^\ddagger$ & \textbf{92.9}$^\ddagger$ & \textbf{78.8}$^\ddagger$ & \textbf{88.5}$^\ddagger$ \\
       \bottomrule
	\end{tabular}
   \caption{Evaluation of our parser when {\em not}~using lemma embeddings (for a more direct comparison with Freda3), on in-domain test datasets.
	 $\ddagger$ of +{\em RL} models represents that the scores are statistically significant at $p<10^{-3}$ with their non-{\em RL} counterparts.
   }
   \label{table:lemma}
   \vspace{-1em}
\end{center}
\end{table}

\paragraph{Evaluating Our Parser without Lemma} Since our baseline \cite{haopeng2017} does not rely on neither lemma or any syntactic information,
we also make a comparison of {\em IPS+ML} and {\em IPS+ML+RL} trained with word and POS embeddings, but without lemma embeddings. The results are given in Table \ref{table:lemma}. We see that our model is still better on average and achieves better performance on all three formalisms. We also notice that the lemma information does not improve the performance in the PAS formalism.

\paragraph{Effect of Reinforcement Learning} 
Fig.~\ref{fig:distance} shows the distributions of the length of the created arcs in the first, second, third and fourth transitions for all words, in the various IPS models in the development corpus.
These distributions show the length of the arcs the models tend to create in the first and later transitions.
Since long arcs are harder to predict, an easy-first strategy would typically amount to creating short arcs first.

In supervised learning ({\em IPS+ML}), there is a slight tendency to create shorter arcs first, but while the ordering is relatively consistent, the differences are small. 
This is in sharp contrast with the distributions we see for our policy gradient parser ({\em IPS+ML+RL}). Here, across the board, it is very likely that the first transition connects neighboring words; and very unlikely that neighboring words are connected at later stages. This suggests that reinforcement learning learns an easy-first strategy of predicting short arcs first. 
Note that unlike easy-first algorithms in syntactic parsing \cite{goldberg2013}, we do not hardwire an easy-first strategy into our parser; but rather, we learn it from the data, because it optimizes our long-term rewards. 
We present further analyses and analyses on WSJ syntactic dependency trees in Appendix~\ref{sec:wsj_dep}.

Fig.~\ref{fig:examples} shows four sentence excerpts from the development corpus, and the order in which arcs are created.
We again compare the model trained with supervised learning ({\em IPS+ML} notated as {\usefont{T1}{pcr}{m}{n} SL} here) to the model with reinforcement learning ({\em IPS+ML+RL} notated as {\usefont{T1}{pcr}{m}{n} RL} here).
In examples (a) and (b), the {\usefont{T1}{pcr}{m}{n} RL} model creates arcs inside noun phrases first and then creates arcs to the verb. The {\usefont{T1}{pcr}{m}{n} SL} model, in contrast, creates arcs with inconsistent orders.
There are lots of similar examples  in the development data.
In clause (c), for example, it seems that the {\usefont{T1}{pcr}{m}{n} RL} model follows a grammatical ordering, while the {\usefont{T1}{pcr}{m}{n} SL} model does not.
In the clause (d), it seems that the {\usefont{T1}{pcr}{m}{n} RL} model first resolves arcs from modifiers, in ``\textit{chief financial officer}'', then creates an arc from the adjective phrase ``\textit{, who will be hired}'', and finally creates an arc from the external phrase ``\textit{the position of}''.
Note that both the {\usefont{T1}{pcr}{m}{n} SL} and {\usefont{T1}{pcr}{m}{n} RL} models make an arc from ``\textit{of}'' in stead of the annotated label of the word ``\textit{position}'' in the phrase ``\textit{the position of}''.
In the clause (e), the {\usefont{T1}{pcr}{m}{n} RL} model resolve the arcs in the noun phrase ``\textit{a strong sales performance}'' and then resolve arcs from the following prepositional phrase. Finally, the {\usefont{T1}{pcr}{m}{n} RL} model resolve the arc from the word ``\textit{with}'' that is the headword in the syntactic dependency tree. In the example (d) and (e), the {\usefont{T1}{pcr}{m}{n} RL} model elaborately follows the syntactic order that are not given in any stages of training and parsing.

\section{Conclusion}

We propose a novel iterative predicate selection (IPS) parsing model for semantic dependency parsing.
We apply multi-task learning to learn general representations of parser configurations, and use reinforcement learning for task-specific fine-tuning.
In our experiments, our multi-task reinforcement IPS model achieves a new state of the art for three SDP formalisms. Moreover, we show that fine-tuning with reinforcement learning learns an easy-first strategy and some syntactic features.

\section*{Acknowledgements}
This work was done when Shuhei Kurita visited the University of Copenhagen.
Shuhei Kurita was supported by JST ACT-I Grant Number JPMJPR17U8, Japan and
partly supported by JST CREST Grant Number JPMJCR1301, Japan.
Anders S{\o}gaard was supported by a Google Focused Research Award.

\bibliography{kurita_acl2019}

\begin{thebibliography}{44}
\expandafter\ifx\csname natexlab\endcsname\relax\def\natexlab#1{#1}\fi

\bibitem[{Almeida and Martins(2015)}]{mariana2015}
M.~Almeida and A.~Martins. 2015.
\newblock Lisbon: Evaluating turbosemanticparser on multiple languages and
  out-of-domain data.

\bibitem[{Baird~III(1999)}]{leemon93}
Leemon~C. Baird~III. 1999.
\newblock Reinforcement learning through gradient descent.
\newblock School of Computer Science Carnegie Mellon University.

\bibitem[{Ballesteros et~al.(2015)Ballesteros, Dyer, and
  Smith}]{ballesteros2015}
Miguel Ballesteros, Chris Dyer, and Noah~A. Smith. 2015.
\newblock \href {http://aclweb.org/anthology/D15-1041} {Improved
  transition-based parsing by modeling characters instead of words with lstms}.
\newblock In \emph{Proceedings of the EMNLP}, pages 349--359.

\bibitem[{Camerini et~al.(1980)Camerini, Fratta, and Maffioli}]{dag_form1980}
P.~M. Camerini, L.~Fratta, and F.~Maffioli. 1980.
\newblock The $k$ best spanning arborescences of a network.
\newblock \emph{Networks}, 10:91--110.

\bibitem[{Copestake et~al.(2005)Copestake, Flickinger, Sag, and
  Pollard}]{copestake2005}
Ann Copestake, Dan Flickinger, Ivan~A. Sag, and Carl Pollard. 2005.
\newblock Minimal recursion semantics: An in- troduction.
\newblock In \emph{Research on Language \& Computation}, pages
  3(4):281–--332.

\bibitem[{Dozat and Manning(2018)}]{dozat2018}
Timothy Dozat and Christopher~D. Manning. 2018.
\newblock Simpler but more accurate semantic dependency parsing.
\newblock In \emph{Proceedings of the ACL (Short Papers)}, pages 484--490.

\bibitem[{Du et~al.(2015)Du, Zhang, Zhang, Sun, and XiaojunWan}]{yantao2015}
Yantao Du, Fan Zhang, Xun Zhang, Weiwei Sun, and XiaojunWan. 2015.
\newblock Peking: Building semantic de- pendency graphs with a hybrid parser.

\bibitem[{Dyer et~al.(2016)Dyer, Kuncoro, Ballesteros, and Smith}]{dyer2016}
Chris Dyer, Adhiguna Kuncoro, Miguel Ballesteros, and Noah~A. Smith. 2016.
\newblock Recurrent neural network grammars.
\newblock In \emph{Proceedings of the 2016 Conference of the NAACL: HLT}, pages
  199--209, San Diego, California.

\bibitem[{Eisner(1996)}]{eisner1996}
J.~Eisner. 1996.
\newblock Three new probabilistic models for dependency parsing: An
  exploration.
\newblock In \emph{COLING}.

\bibitem[{Flickinger et~al.(2012)Flickinger, Zhang, and Kordoni}]{deepbank}
Daniel Flickinger, Yi~Zhang, and Valia Kordoni. 2012.
\newblock Deepbank: Adynamically annotated treebank of the wall street journal.
\newblock In \emph{In Proc. of TLT}.

\bibitem[{Fried and Klein(2018)}]{daniel2018}
Daniel Fried and Dan Klein. 2018.
\newblock Policy gradient as a proxy for dynamic oracles in constituency
  parsing.
\newblock In \emph{Proceedings of the ACL}, pages 469--476.

\bibitem[{Galley and Manning(2009)}]{galley2009}
Michel Galley and Christopher~D. Manning. 2009.
\newblock Quadratic-time dependency parsing for machine translation.
\newblock In \emph{Proceedings of the Joint Conference of the 47th Annual
  Meeting of the ACL and the 4th International Joint Conference on Natural
  Language Processing of the AFNLP}, pages 773--781. Association for
  Computational Linguistics.

\bibitem[{Goldberg and Elhadad(2010)}]{goldberg2010}
Yoav Goldberg and Michael Elhadad. 2010.
\newblock An efficient algorithm for easy-first non-directional dependency
  parsing.
\newblock In \emph{Human Language Technologies: NAACL}, pages 742--750, Los
  Angeles, California.

\bibitem[{Goldberg and Nivre(2013)}]{goldberg2013}
Yoav Goldberg and Joakim Nivre. 2013.
\newblock Training deterministic parsers with non-deterministic oracles.
\newblock pages 403--414.

\bibitem[{Haji\v{c} et~al.(2012)Haji\v{c}, Haji\v{c}ov{\'a}, Panevov{\'a},
  Sgall, Bojar, Cinkov{\'a}, Fu\v{c}{\'\i}kov{\'a}, Mikulov{\'a}, Pajas,
  Popelka, Semeck{\'y}, \v{S}indlerov{\'a}, \v{S}t\v{e}p{\'a}nek, Toman,
  Ure\v{s}ov{\'a}, and \v{Z}abokrtsk{\'y}}]{HAJI12.510.L12-1280}
Jan Haji\v{c}, Eva Haji\v{c}ov{\'a}, Jarmila Panevov{\'a}, Petr Sgall,
  Ond\v{r}ej Bojar, Silvie Cinkov{\'a}, Eva Fu\v{c}{\'\i}kov{\'a}, Marie
  Mikulov{\'a}, Petr Pajas, Jan Popelka, Ji\v{r}{\'\i} Semeck{\'y}, Jana
  \v{S}indlerov{\'a}, Jan \v{S}t\v{e}p{\'a}nek, Josef Toman, Zde\v{n}ka
  Ure\v{s}ov{\'a}, and Zden\v{e}k \v{Z}abokrtsk{\'y}. 2012.
\newblock Announcing prague czech-english dependency treebank 2.0.
\newblock In \emph{Proceedings of the Eighth International Conference on
  Language Resources and Evaluation (LREC-2012)}, pages 3153--3160.

\bibitem[{Hashimoto et~al.(2017)Hashimoto, Xiong, Tsuruoka, and
  Socher}]{hashimoto2017}
Kazuma Hashimoto, Caiming Xiong, Yoshimasa Tsuruoka, and Richard Socher. 2017.
\newblock \href {http://aclweb.org/anthology/D17-1206} {A joint many-task
  model: Growing a neural network for multiple nlp tasks}.
\newblock In \emph{Proceedings of the EMNLP}, pages 1923--1933.

\bibitem[{Hershcovich et~al.(2017)Hershcovich, Abend, and
  Rappoport}]{hershcovich2017}
Daniel Hershcovich, Omri Abend, and Ari Rappoport. 2017.
\newblock A transition-based directed acyclic graph parser for ucca.
\newblock In \emph{Proceedings of the 55th Annual Meeting of the Association
  for Computational Linguistics (Volume 1: Long Papers)}, pages 1127--1138.
  Association for Computational Linguistics.

\bibitem[{Kiperwasser and Goldberg(2016)}]{kiperwasser2016}
Eliyahu Kiperwasser and Yoav Goldberg. 2016.
\newblock \href {https://transacl.org/ojs/index.php/tacl/article/view/885}
  {Simple and accurate dependency parsing using bidirectional lstm feature
  representations}.
\newblock \emph{TACL}, 4:313--327.

\bibitem[{Lee et~al.(2018)Lee, Mansimov, and Cho}]{jason2018}
Jason Lee, Elman Mansimov, and Kyunghyun Cho. 2018.
\newblock Deterministic non-autoregressive neural sequence modeling by
  iterative refinement.
\newblock In \emph{Proceedings of the 2018 Conference on Empirical Methods in
  Natural Language Processing}, pages 1173--1182. Association for Computational
  Linguistics.

\bibitem[{Ma et~al.(2013)Ma, Zhu, Xiao, and Yang}]{ma-etal-2013-easy}
Ji~Ma, Jingbo Zhu, Tong Xiao, and Nan Yang. 2013.
\newblock \href {https://www.aclweb.org/anthology/P13-2020} {Easy-first {POS}
  tagging and dependency parsing with beam search}.
\newblock In \emph{Proceedings of the 51st Annual Meeting of the Association
  for Computational Linguistics (Volume 2: Short Papers)}, pages 110--114,
  Sofia, Bulgaria. Association for Computational Linguistics.

\bibitem[{Martins et~al.(2011)Martins, Smith, Figueiredo, and
  Aguiar}]{martins-EtAl:2011:EMNLP1}
Andre Martins, Noah Smith, Mario Figueiredo, and Pedro Aguiar. 2011.
\newblock Dual decomposition with many overlapping components.
\newblock In \emph{Proceedings of the 2011 Conference on EMNLP}, pages
  238--249, Edinburgh, Scotland, UK.

\bibitem[{Martins and Kreutzer(2017)}]{martins-kreutzer-2017-learning}
Andr{\'e} F.~T. Martins and Julia Kreutzer. 2017.
\newblock \href {https://doi.org/10.18653/v1/D17-1036} {Learning what{'}s easy:
  Fully differentiable neural easy-first taggers}.
\newblock In \emph{Proceedings of the 2017 Conference on Empirical Methods in
  Natural Language Processing}, pages 349--362, Copenhagen, Denmark.
  Association for Computational Linguistics.

\bibitem[{McDonald and Nivre(2007)}]{McDonald:Nivre:07}
Ryan McDonald and Joakim Nivre. 2007.
\newblock Characterizing the errors of data-driven dependency parsing models.
\newblock In \emph{Proceedings of the 2007 Joint Conference on EMNLP-CoNLL},
  pages 122--131.

\bibitem[{McDonald and Pereira(2006)}]{mcdonald2006}
Ryan McDonald and Fernando Pereira. 2006.
\newblock \href {http://aclweb.org/anthology/E06-1011} {Online learning of
  approximate dependency parsing algorithms}.
\newblock In \emph{11th Conference of the European Chapter of the Association
  for Computational Linguistics}.

\bibitem[{Miyao et~al.(2004)Miyao, Ninomiya, and Tsujii}]{miyao2004}
Yusuke Miyao, Takashi Ninomiya, and Jun'ichi. Tsujii. 2004.
\newblock Corpus-oriented grammar development for acquiring a head-driven
  phrase structure grammar from the penn treebank.
\newblock In \emph{In Proceedings of IJCNLP-04}.

\bibitem[{Nivre and Scholz(2004b)}]{nivre2004arceager}
Joakim Nivre and Mario Scholz. 2004b.
\newblock Deterministic dependency parsing of english text.
\newblock In \emph{Proceedings of Coling 2004}, pages 64--70. COLING.

\bibitem[{Oepen et~al.(2015)Oepen, Kuhlmann, Miyao, Zeman, Cinkova, Flickinger,
  Hajic, and Uresova}]{oepen-EtAl:2015:SemEval}
Stephan Oepen, Marco Kuhlmann, Yusuke Miyao, Daniel Zeman, Silvie Cinkova, Dan
  Flickinger, Jan Hajic, and Zdenka Uresova. 2015.
\newblock \href {http://www.aclweb.org/anthology/S15-2153} {Semeval 2015 task
  18: Broad-coverage semantic dependency parsing}.
\newblock In \emph{Proceedings of the 9th International Workshop on Semantic
  Evaluation (SemEval 2015)}, pages 915--926, Denver, Colorado. Association for
  Computational Linguistics.

\bibitem[{Oepen et~al.(2014)Oepen, Kuhlmann, Miyao, Zeman, Flickinger, Hajic,
  Ivanova, and Zhang}]{oepen-EtAl:2014:SemEval}
Stephan Oepen, Marco Kuhlmann, Yusuke Miyao, Daniel Zeman, Dan Flickinger, Jan
  Hajic, Angelina Ivanova, and Yi~Zhang. 2014.
\newblock \href {http://www.aclweb.org/anthology/S14-2008} {Semeval 2014 task
  8: Broad-coverage semantic dependency parsing}.
\newblock In \emph{Proceedings of the 8th International Workshop on Semantic
  Evaluation (SemEval 2014)}, pages 63--72, Dublin, Ireland.

\bibitem[{Peng et~al.(2017)Peng, Thomson, and Smith}]{haopeng2017}
Hao Peng, Sam Thomson, and Noah~A. Smith. 2017.
\newblock Deep multitask learning for semantic dependency parsing.
\newblock In \emph{Proceedings of the ACL}, pages 2037--2048, Vancouver,
  Canada.

\bibitem[{Peng et~al.(2018{\natexlab{a}})Peng, Thomson, and
  Smith}]{haopeng2018b}
Hao Peng, Sam Thomson, and Noah~A. Smith. 2018{\natexlab{a}}.
\newblock Backpropagating through structured argmax using a spigot.
\newblock In \emph{Proceedings of the 56th Annual Meeting of the ACL}, pages
  1863--1873.

\bibitem[{Peng et~al.(2018{\natexlab{b}})Peng, Thomson, Swayamdipta, and
  Smith}]{peng-etal-2018-learning}
Hao Peng, Sam Thomson, Swabha Swayamdipta, and Noah~A. Smith.
  2018{\natexlab{b}}.
\newblock \href {https://doi.org/10.18653/v1/N18-1135} {Learning joint semantic
  parsers from disjoint data}.
\newblock In \emph{Proceedings of the 2018 Conference of the North {A}merican
  Chapter of the Association for Computational Linguistics: Human Language
  Technologies, Volume 1 (Long Papers)}, pages 1492--1502, New Orleans,
  Louisiana. Association for Computational Linguistics.

\bibitem[{Pennington et~al.(2014)Pennington, Socher, and
  Manning}]{pennington2014glove}
Jeffrey Pennington, Richard Socher, and Christopher~D Manning. 2014.
\newblock Glove: Global vectors for word representation.
\newblock In \emph{EMNLP}, volume~14, pages 1532--1543.

\bibitem[{Ribeyre et~al.(2014)Ribeyre, Villemonte de~la Clergerie, and
  Seddah}]{ribeyre-villemontedelaclergerie-seddah:2014:SemEval}
Corentin Ribeyre, Eric Villemonte de~la Clergerie, and Djam\'{e} Seddah. 2014.
\newblock Alpage: Transition-based semantic graph parsing with syntactic
  features.
\newblock In \emph{Proceedings of the 8th International Workshop on Semantic
  Evaluation (SemEval 2014)}, pages 97--103, Dublin, Ireland. Association for
  Computational Linguistics and Dublin City University.

\bibitem[{Sagae and Tsujii(2008)}]{sagae2008}
Kenji Sagae and Jun'ichi Tsujii. 2008.
\newblock Shift-reduce dependency {DAG} parsing.
\newblock In \emph{Proceedings of the 22nd International Conference on
  Computational Linguistics (Coling 2008)}, pages 753--760. Coling 2008
  Organizing Committee.

\bibitem[{S{\o}gaard and Goldberg(2016)}]{ander2016}
Anders S{\o}gaard and Yoav Goldberg. 2016.
\newblock \href {https://doi.org/10.18653/v1/P16-2038} {Deep multi-task
  learning with low level tasks supervised at lower layers}.
\newblock In \emph{Proceedings of the ACL (Short Papers)}, pages 231--235.

\bibitem[{Spitkovsky et~al.(2011)Spitkovsky, Alshawi, Chang, and
  Jurafsky}]{spitkovsky2011}
Valentin~I. Spitkovsky, Hiyan Alshawi, Angel~X. Chang, and Daniel Jurafsky.
  2011.
\newblock \href {http://www.aclweb.org/anthology/D11-1118} {Unsupervised
  dependency parsing without gold part-of-speech tags}.
\newblock In \emph{Proceedings of the 2011 Conference on EMNLP}, pages
  1281--1290.

\bibitem[{Tokg\"{o}z and G\"{u}lsen(2015)}]{alper2015}
Alper Tokg\"{o}z and Eryigit G\"{u}lsen. 2015.
\newblock Transition-based dependency dag parsing using dynamic oracles.
\newblock In \emph{Proceedings of the ACL Student Research Workshop.}, pages
  22--27.

\bibitem[{Tsuruoka and Tsujii(2005)}]{tsuruoka-tsujii-2005-bidirectional}
Yoshimasa Tsuruoka and Jun{'}ichi Tsujii. 2005.
\newblock \href {https://www.aclweb.org/anthology/H05-1059} {Bidirectional
  inference with the easiest-first strategy for tagging sequence data}.
\newblock In \emph{Proceedings of Human Language Technology Conference and
  Conference on Empirical Methods in Natural Language Processing}, pages
  467--474, Vancouver, British Columbia, Canada. Association for Computational
  Linguistics.

\bibitem[{Wang et~al.(2018)Wang, Che, Guo, and Liu}]{wang2018}
Yuxuan Wang, Wanxiang Che, Jiang Guo, and Ting Liu. 2018.
\newblock A neural transition-based approach for semantic dependency graph
  parsing.
\newblock In \emph{Proceedings of the Thirty-Second {AAAI} Conference on
  Artificial Intelligence}.

\bibitem[{Williams(1992)}]{policygradient1992}
Ronald~J Williams. 1992.
\newblock Simple statistical gradient-following algorithms for connectionist
  reinforcement learning.
\newblock pages 5--32. Springer.

\bibitem[{Xu et~al.(2009)Xu, Kang, Ringgaard, and Och}]{Xu:ea:09}
Peng Xu, Jaeho Kang, Michael Ringgaard, and Franz Och. 2009.
\newblock \href {http://www.aclweb.org/anthology/N/N09/N09-1028} {Using a
  dependency parser to improve smt for subject-object-verb languages}.
\newblock In \emph{Proceedings of HLT:NAACL}, pages 245--253, Boulder,
  Colorado.

\bibitem[{Zhang and Chan(2009)}]{zhangchan2009}
Lidan Zhang and Kwok~Ping Chan. 2009.
\newblock \href {http://www.aclweb.org/anthology/W09-3838} {Dependency parsing
  with energy-based reinforcement learning}.
\newblock In \emph{Proceedings of the IWPT}, pages 234--237, Paris, France.

\bibitem[{Zhang et~al.(2017)Zhang, Cheng, and Lapata}]{zhang-cheng-lapata2017}
Xingxing Zhang, Jianpeng Cheng, and Mirella Lapata. 2017.
\newblock Dependency parsing as head selection.
\newblock In \emph{Proceedings of the ACL}, pages 665--676, Valencia, Spain.

\bibitem[{Zhang and Clark(2008)}]{zhang-clark2008:EMNLP}
Yue Zhang and Stephen Clark. 2008.
\newblock A tale of two parsers: {I}nvestigating and combining graph-based and
  transition-based dependency parsing.
\newblock In \emph{Proceedings of the EMNLP}, pages 562--571.

\end{thebibliography}
\bibliographystyle{acl_natbib}

\clearpage
\appendix

\section{Appendix}

\begin{table}[H]
\begin{center}
	\small\begin{tabular}{lccc}
        \toprule
        Model               & DM (\%)  & PAS (\%)  & PSD (\%) \\
        \midrule
	    IPS +ML		  & 27 (1.6\%) & 15 (0.88\%) & 25 (1.4\%) \\
	    IPS +ML +RL	  & 25 (1.5\%) & 19 (1.1\%) & 13 (0.76\%) \\
        \bottomrule
	\end{tabular}
    \caption{
        The number of graphs and the percentages of graphs with circles when not using heuristics to avoid circles. Results on the development sets. We also note circles are mostly small and local. So they do not affect other arc structures.
    }
    \vspace{-0.5em}
    \label{table:dag_stat}
\end{center}
\end{table}

\subsection{DAG formalism}
\label{sec:dag_stat}

Our proposed parsing algorithm possibly introduces circles in the resulting graphs. However, they are few in our experiments. Table~\ref{table:dag_stat} shows the number of graphs with circles and their relative frequency in our predictions for the development sets. Here we propose additional decoding rules to strictly prevent making these circles.
Our decoding is iterative. Therefore when the newly arcs $A$ are added to the existing partial SDP graph $\mathbf{y}^{\tau}$, the sum of $\mathbf{y}^{\tau} \cup A$ must not contain some circles. In that case, we swap  arcs in $A$ with other arcs that do not lead to circles, in the following way:
\begin{description}
	\item[1] Search for arcs that make circles $C$ in $\mathbf{y}^{\tau} \cup A$. If no circles (most in the case), return $\mathbf{y}^{\tau} \cup A$ as a partial SDP graph and go to the next transition.
	\item[2] Prepare buffers for reduced arcs $B(={\emptyset})$ and for alternative arcs $B'(={\emptyset})$.
    \item[3] From arcs in $C \cap A \backslash B$, choose one arc $a$ that have the lowest probability of the softmax output and add to the buffer $B$.
    \item[4] For the reduced arc $a$, choose another arc (including {\usefont{T1}{pcr}{m}{n} NULL}) that have the next largest probability of the softmax output of the arc $a$ and add it to the buffer $B'$.
	\item[5] Check if $\mathbf{y}^{\tau} \cup A \backslash B$ have circles. If no circles, add $B'$ to $A$ and go to {\bf 1}. Otherwise, go to {\bf 3}.
\end{description}
The resulting graphs do not contain circles and form DAGs.
This operation slightly improves the performance, but does not affect the results much, because circles are rare.

\subsection{Arc Length Analysis on WSJ Syntactic Trees}
\label{sec:wsj_dep}

\begin{figure}[t]
	\centering
	\includegraphics[scale=0.6,clip]{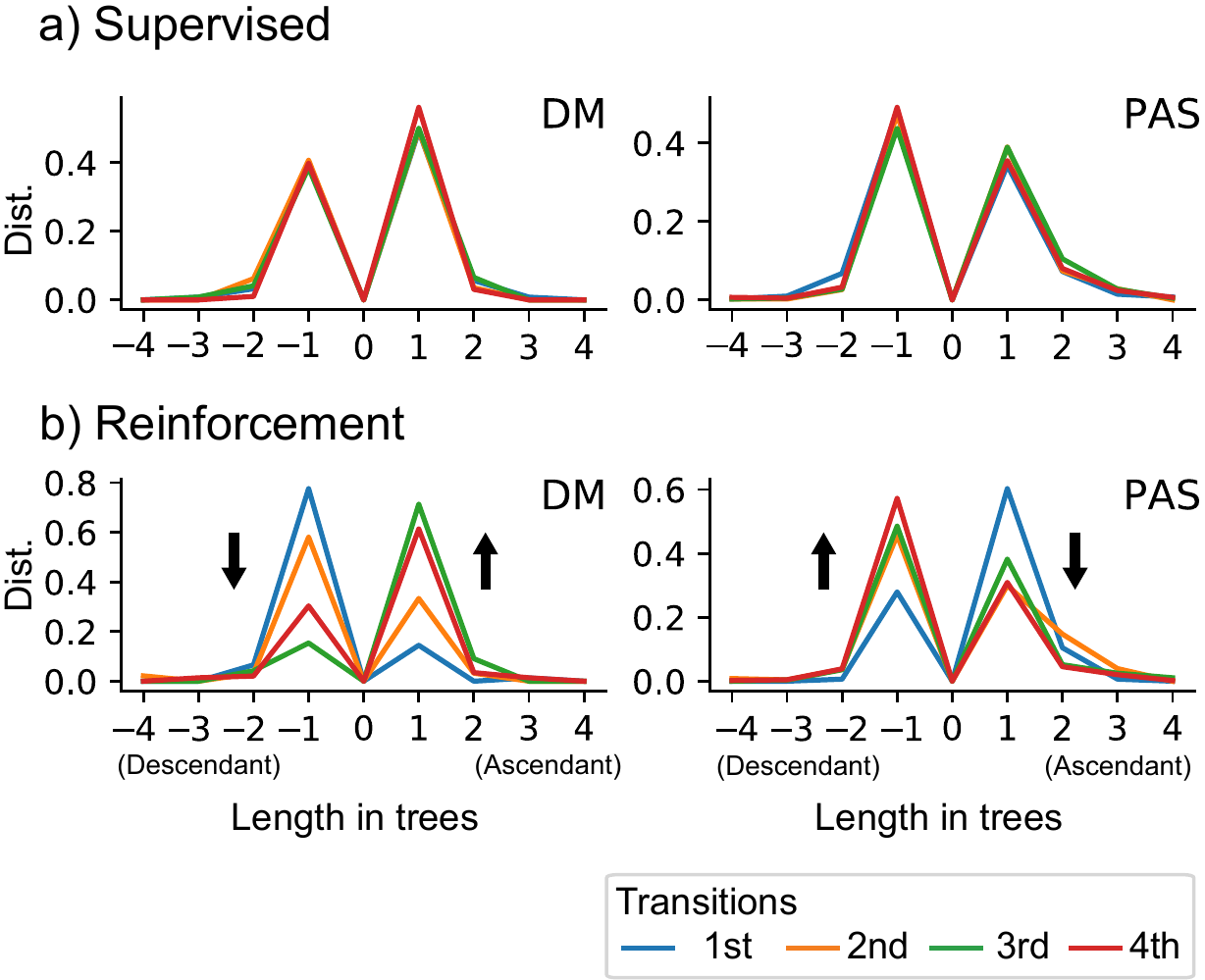}
	\vspace{-0.5em}
       \caption{
			 Arc length distributions on the WSJ syntactic dependency trees.
			 (a) The length distribution of arcs with supervised learning ({\em IPS+ML}).
			 (b) The length distribution of arcs with reinforcement learning  ({\em IPS+ML+RL}).
			 The four lines correspond to the first four transitions in the derivations.
			 The horizontal axis corresponds to the length of the created arcs. The rightward of the horizontal axis corresponds to arcs from ascendants, while the leftward corresponds to arcs from descendants.
			 The black arrows in the bottom figures illustrate the changes of distributions from the 1st to later distributions. They are in the opposite directions between DM and PAS.
       }
       \vspace{-0.5em}
       \label{fig:distance_on_dep}
\end{figure}

\begin{figure}[t]
	\centering
	\includegraphics[scale=0.6,clip]{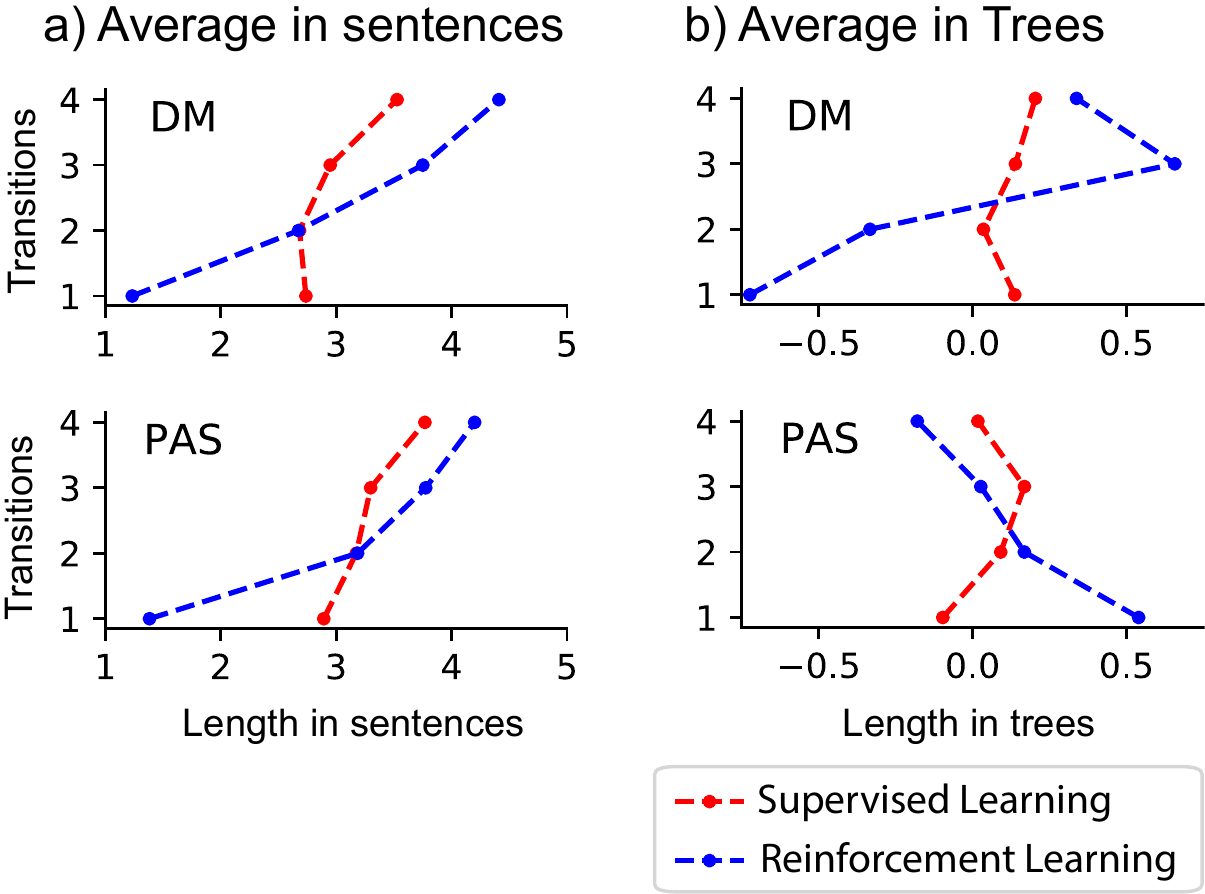}
	\vspace{-0.5em}
       \caption{
			 The average arc length comparisons of the supervised learning model ({\em IPS+ML}) and the reinforcement learning model ({\em IPS+ML+RL}).
			 (a) Average arc length in sentences. The horizontal axis is the length of arcs in terms of words as of Fig. \ref{fig:distance}. The vertical axis corresponds to the first four transitions of models.
       (b) Average arc length in dependency trees. The horizontal axis is the same with Fig.~\ref{fig:distance_on_dep}.
       }
       \vspace{-0.5em}
       \label{fig:distance_averaged}
\end{figure}

\begin{figure}[t]
	\centering
	\includegraphics[scale=0.85,clip]{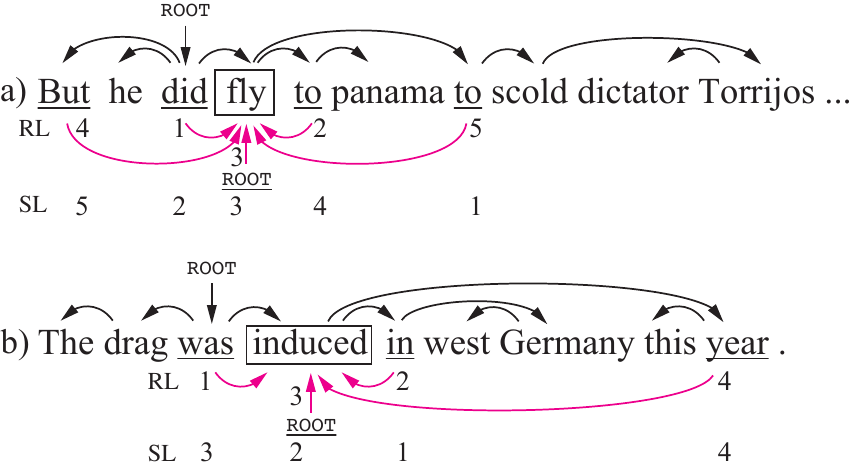}
	\vspace{-0.5em}
       \caption{Examples of clauses parsed with PAS formalism. We present the gold syntactic trees above the sentences while the PAS arcs to a word under the sentence. We also attach the numbers of parsing orders. RL represents the parsing orders of the reinforcement learning model ({\it IPS+ML+RL}) while SL represents those of supervised leaning model({\it IPS+ML}) on the PAS graphs.
       }
       \vspace{-0.5em}
       \label{fig:error_pas}
\end{figure}

As a supplementary experiment, we wanted to explore the relation between the learned easy-first strategies and syntactic dependency lengths. The intuition was that attachments of length $n$ that are consistent with syntactic dependencies may be easier than attachments of length $n$ that are {\em not} consistent with syntactic dependencies, regardless of $n$. We therefore provide a quantitative analysis of the SDP parsing order using syntactic dependency trees of WSJ corpus as a reference.
The syntactic dependency trees are extracted from WSJ constituency trees with the LTH Penn Converter.\footnote{\url{http://nlp.cs.lth.se/software/treebank_converter/}}\footnote{We note that we use these syntactic trees only for this supplementary analysis; and {\em not} for training or validating semantic parsers.} 
For this, we consider the subset of undirected SDP arcs that match syntactic dependencies, i.e., where the semantic predicate words and the semantic argument words are in ascendant or descendant relations in the syntactic dependency trees.
The directed SDP arcs that agree with the syntactic dependencies in direction, i.e., such that the syntactic head is also the semantic predicate, are said to have positive answer, as in the analysis above, while SDP arcs that are opposite have negative distance, because they go against the syntactic order.
We show the distributions of length of arcs to semantic headwords that are created from the 1st to the 4th transitions from semantic argument words.
We consider words that have four or more semantic headwords; this is because the models finish transitions for words that have fewer semantic headwords in the early transitions.

Fig.~\ref{fig:distance_on_dep} presents the distributions of arc length from the 1st to the 4th transitions. 
These graphs are similar to Fig.~\ref{fig:distance}, but only represent a subset of arcs, and now with a distinction between positive and negative arcs, depending on whether they agree with the syntactic analysis. We present the DM and PAS graphs, because we had too few examples for PSD. The graph (a) of supervised learning shows the same tendency from 1st to 4th transitions. The graph (b) of reinforcement learning shows a different picture: For PAS, the {\it IPS+ML+RL} model tends to resolve the short arcs that are consistent with the syntactic ones first (the blue line is higher in the positive span). 
In DM, however, the model tends to resolve the short arcs that are {\em inconsistent}~with syntactic dependencies first. 

Fig.~\ref{fig:distance_averaged} presents the averaged arc lengths at each transition step. The left column (a) is the average length of the created arcs at transition steps one to four. The right column (b) is the same relative to syntactic trees, using the same subsample as above and encoding disagreeing arcs with negative length values. In supervised learning, the averaged arc length does not vary much across transition steps, neither with respect to sentence positions nor with respect to syntactic trees. However, in reinforcement learning, the averaged arc length varies a lot across transition steps, relative to both sentence positions and agreement with syntactic headedness. 
The graphs in the (a) column suggest that the reinforcement learning model has a strong tendency to resolve adjacent arcs first. In the (b) column, we note that for reinforcement learning, the trajectories for DM and PAS are opposite. For PAS, early arcs are in line with syntactic dependencies, whereas for DM the opposite picture emerges.

Fig.~\ref{fig:error_pas} presents two example clauses with gold syntactic trees the partial SDP graphs, decorated with the parsing orders of our supervised baseline model (SL) and our reinforcement learning model (RL). RL resolves the adjacent words first, and the parsing orders of clauses (a) and (b) are consistent. The two PAS graphs and syntactic trees are relatively similar, but the arc directions are different, and the arcs from {\usefont{T1}{pcr}{m}{n} ROOT} go to different words. The RL model prefers to create arcs that agree with the syntactic arcs first. We also note that the parsing orders of the SL models seem inconsistent across (a) and (b). 





\end{document}